\pdfoutput=1

\documentclass[11pt]{article}

\usepackage[final]{acl}

\usepackage{times}
\usepackage{latexsym}

\usepackage[T1]{fontenc}

\usepackage[utf8]{inputenc}

\usepackage{microtype}

\usepackage{multirow}
\usepackage{graphicx}
\usepackage{balance}
\usepackage{xcolor}
\usepackage{hyperref}

\title{Teaching Small Language Models to Reason}

\author{%
Lucie Charlotte Magister\thanks{~~Research conducted during an internship at Google.} \\
  University of Cambridge \\
  \texttt{lcm67@cam.ac.uk}\\\And
  Jonathan Mallinson \\
  Google Research \\
  \texttt{jonmall@google.com}\\\And
  Jakub Adamek \\
  Google Research \\
  \texttt{enkait@google.com}\\\AND
  Eric Malmi \\
  Google Research \\
  \texttt{emalmi@google.com}\\\And
  Aliaksei Severyn \\
  Google Research \\
  \texttt{severyn@google.com} \\}
  
\begin{document}
\maketitle

\begin{abstract}

Chain of thought prompting successfully improves the reasoning capabilities of large language models, achieving state of the art results on a range of datasets. However, these reasoning capabilities only appear to emerge in models with at least tens of billions of parameters. In this paper, we explore the transfer of such reasoning capabilities to smaller models via knowledge distillation, also investigating model and dataset size trade-off. Specifically, we finetune a student model on the chain of thought outputs generated by a larger teacher model. Our experiments show that the proposed method improves task performance across arithmetic, commonsense and symbolic reasoning datasets. For example, the accuracy of T5 XXL on GSM8K improves from 8.11\% to 21.99\% and 18.42\% when finetuned on PaLM 540B and GPT-3 175B generated chains of thought, respectively.

\end{abstract}

\section{Introduction}

Chain of thought (CoT) prompting encourages language models (LMs) to break down a reasoning task into a series of intermediate steps \citep{wei2022chain}. They demonstrate that this prompting significantly increases the task accuracy of large language models (LLMs) across commonsense, symbolic and mathematical reasoning datasets. Here, LLMs are models with at least tens of billions of parameters, such as PaLM 540B \citep{chowdhery2022palm}, GPT-3 175B \citep{brown2020language}, or UL2 20B \citep{tay2022unifying}. However, the reasoning capabilities of smaller LMs do not improve with CoT prompting, mostly producing illogical CoT. Notably, CoT prompting even reduces the accuracy of models with less than 10 billion parameters. \citet{wei2022chain} attribute this to abilities, such as semantic understanding and symbolic mapping, only emerging at larger scales. This leads us to our research question: \emph{can the reasoning capabilities of LLMs be transferred to smaller LMs via finetuning?}

This work explores CoT knowledge distillation \cite{hinton2015distilling} from PaLM 540B \citep{chowdhery2022palm} and GPT-3 175B \citep{brown2020language} to different sizes of the smaller language model T5~\citep{raffel2020exploring}, such as T5 XXL, XL and base, which have 11 billion, 3 billion and 220 million parameters, respectively. As a result of our work, we make two recommendations: (1) perform knowledge distillation by finetuning the student model on the CoT generated by a large teacher model; and (2) generate the CoT from an LLM, as proposed by \citet{wei2022chain}, but crucially provide the solution to the task in the few-shot prompt. We demonstrate that the proposed method improves task performance across arithmetic, commonsense and symbolic reasoning datasets irrespective of the teacher model used. For example, we show an accuracy increase from 8.11\% to 21.99\% and 18.42\% on the GSM8K \citep{cobbe2021training} dataset when finetuning T5 XXL on PaLM 540B and GPT-3 175B generated CoT data, respectively.

\section{Related Work}
This work is inspired by the seminal work of \citet{wei2022chain} on CoT prompting. They demonstrate that prefixing an input with 2-8 exemplars of CoT reasoning encourages LMs to do the same, reaching state-of-the-art performance on datasets such as GSM8K \citep{cobbe2021training}. \citet{wang2022self} show that task accuracy can be further improved by using self-consistency in CoT prompting. Self-consistency samples CoT reasoning paths from a model's decoder and returns the most consistent path by taking the majority vote. Subsequently, \citet{chung2022scaling} explore finetuning a FLAN-based \citep{wei2021finetuned} version of PaLM on manually generated CoT data.

Concurrent to our work, a small number of other works propose methods focused on CoT student--teacher knowledge distillation. \citet{ho2022reasoningteachers} and \citet{li2022explanations} also explore knowledge distillation with the difference of proposing diverse sampling and rationalization prompting, respectively. In contrast to their work, our work explores more teacher models and demonstrates both the effects of dataset and model size on accuracy. We also achieve a higher accuracy on common datasets, such as GSM8K, than \citet{ho2022reasoningteachers}. In contrast to our work, \citet{shridhar2022distilling} focus on training two models, one for problem decomposition and one for solving. Yet differently, the focus of \citet{eisenstein2022honest} relies on producing markup-and-mask explanations for open-book question answering. Lastly, \citet{huang2022large} present one related experiment, however, we present a more in-depth exploration on more datasets. To the best of our knowledge, our work is the first to extensively explore the improvement of the reasoning ability of small LMs via knowledge distillation across multiple model architectures, and observing the effects of student model size and dataset size on accuracy.

\section{Method}
\label{method_section}

\begin{figure}[th]
    \centering
    \includegraphics[width=\columnwidth]{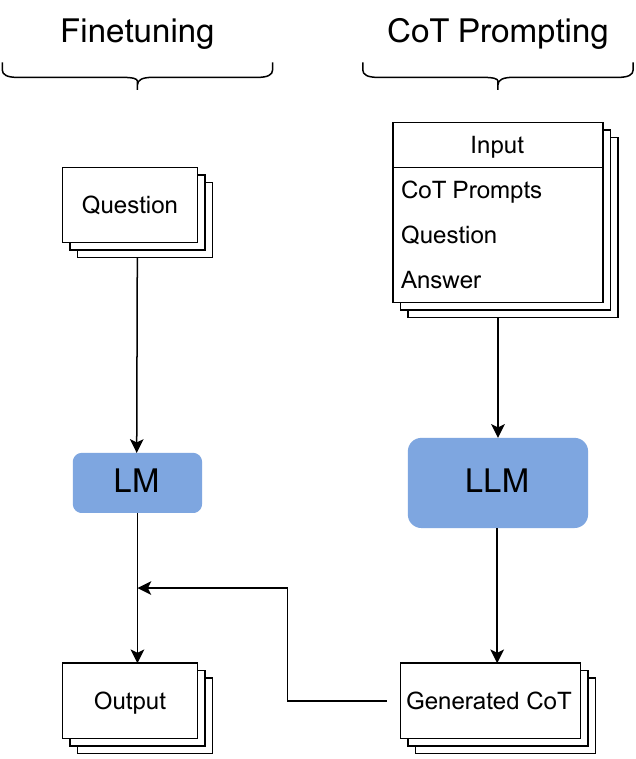}
    \caption{Overview of the proposed method.}
    \label{fig:visual_abstract}
\end{figure}

We propose a two-step pipeline for CoT knowledge distillation. The first step comprises annotating an existing supervised dataset with CoT reasoning generated by a teacher model. To generate high quality data, we propose using LLMs, such as PaLM 540B or GPT-3 175B, as teachers, based on the finding that CoT reasoning improves with model scale \citep{wei2022chain}. Specifically, we perform few-shot prompting with 8 exemplars on these models to generate CoTs. However, we make a key modification to the prompts proposed by \citet{wei2022chain}. We adapt the few-shot prompts to provide the model with the target after posing the question and before providing example CoT. This is based on the observation that providing this guidance allows LLMs to correct small mistakes in the CoT. Lastly, we remove all incorrect CoT based on the target answer to prevent the student to learn from bad examples. The second step comprises finetuning a student model via teacher forcing \citep{williams1989learning}. The student is provided with the question as input, and the CoT and answer as the target. As the model is trained on producing a CoT during finetuning, prompting is not required. Figure \ref{fig:visual_abstract} provides an overview of the proposed method.

\section{Experimental Setup}
\label{experimental_setup}

We follow a similar experimental setup to \citet{wei2022chain}, focusing on tasks covering arithmetic, commonsense and symbolic reasoning. 

\subsection{Benchmarks and Metrics}

\subsubsection{Arithmetic Reasoning}
We benchmark the proposed method on the following math word problem datasets: (1) GSM8K \citep{cobbe2021training}, (2) MAWPS \citep{koncel2016mawps} and (3) ASDiv \citep{miao2021diverse}. We use the official training and testing split for GSM8K, taking the last 10\% of the training split for validation, and the 5-fold cross validation splits available for MAWPS and ASDiv. We evaluate task accuracy by checking for the target answer as the final answer in the CoT. In addition, we compute the task accuracy given an external calculator, to account for arithmetic mistakes made by the model, despite the CoT being correct. The external calculator moves through the generated output, recalculating the left hand-side of equations. It then replaces the right-hand side with the calculated output, to avoid arithmetic mistakes being carried forward. For example, if a model outputted '5 + 5 = 11. 11 * 2 = 22', then the external calculator would first calculate '5+5' and replace the '11' with a '10'. In the subsequent equation, it would also replace the '11' with a '10' and arrive at the final result of '20'.


\subsubsection{Commonsense Reasoning}
We benchmark the model's ability to perform commonsense reasoning on the StrategyQA dataset \citep{geva2021did}. As a testing split is not available, we do not shuffle the dataset to allow reproducing our split of taking the first 80\% as training data, the following 10\% as validation data, and the final 10\% as testing data. We compute task accuracy in the same manner as previously mentioned.


\subsubsection{Symbolic Reasoning}
Lastly, we benchmark the model on two synthetic tasks for symbolic reasoning: (1) last letter concatenation and (2) coinflip \citep{wei2022chain}. Last letter concatenation prompts the model to concatenate the last letter of each word in a string. Coinflip prompts the model to perform state tracking of the coin being flipped. We evaluate task accuracy in the same manner as before. Due to the rigid structure of the datasets, we focus on evaluating the model's generalizability to out-of-distribution (OOD) examples. We finetune the models on examples of length two and evaluate on sequences of length three and four. We initially infer the CoT using PaLM 540B, however, find that the LLM is able to perfectly replicate the desired CoT bar one example due to the rigidness of the template. We therefore decide to use the template generated CoT in our experiments.

\subsection{Baselines and setup}

We select PaLM 540B \citep{chowdhery2022palm} and GPT-3 175B \citep{brown2020language} as teacher models. We select PaLM 540B based on the state-of-the-art results on the benchmarking datasets reported by \citet{wei2022chain}, and confirm the observed trends with GPT-3 175B. The publicly accessible teacher models are prompted as described in Section \ref{method_section}.

\begin{figure}[tb]
    \centering
    \includegraphics[width=0.9\columnwidth]{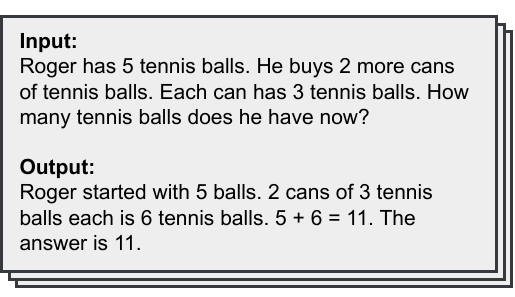}
    \caption{A training example from \citet{wei2022chain} demonstrating the input and output provided to T5.}
    \label{fig:training_example}
\end{figure}

We select different sizes of T5 \citep{raffel2020exploring} as student models, as T5 is publicly available in many sizes. The student models are trained on the PaLM 540B or GPT-3 175B generated CoT data as described in Section \ref{method_section}. We establish T5 XXL model finetuned on the original target as the baseline. We refrain from shuffling the datasets to allow for reproducibility.For the MAWPS and ASDiv dataset, we perform 5-fold cross validation. For all remaining datasets, we take 10\% of the training set as a validation set to select the best model checkpoint. Figure \ref{fig:training_example} showcases an input examples for T5. We refer the reader to \citet{wei2022chain} for more training examples, as well as the prompts used for generating the CoT using PaLM 540B and GPT-3 175B.

We refer the reader to Appendix \ref{licenses} for an overview of the dataset licenses. We also refer the reader to Appendix \ref{compute} for an overview of the computatinal resources.

\section{Results}

\subsection{Arithmetic reasoning}
Table \ref{arithmetic_reasoning_table} details the task accuracy with and without an external calculator for the arithmetic reasoning benchmarks. Our results show that the proposed method improves task accuracy across all datasets. Most notably, the task accuracy of MAWPS is significantly improved. The accuracy achieved given a calculator comes close to the accuracy of 8-shot PaLM 540B, demonstrating that knowledge distillation is effective, but potentially limited by the mathematical abilities of small models.

\begin{table}[ht]
\Huge
\centering
\resizebox{\columnwidth}{!}{%
\begin{tabular}{lccccc}
\hline
\multirow{2}{*}{} & \multicolumn{1}{l}{\textbf{\begin{tabular}[c]{@{}c@{}}Baseline\\ T5 XXL\end{tabular}}} & \multicolumn{2}{c}{\textbf{\begin{tabular}[c]{@{}c@{}}CoT Finetuned \\ T5 XXL\end{tabular}}} &
\multicolumn{2}{c}{\textbf{\begin{tabular}[c]{@{}c@{}}CoT 8-shot \\ PaLM 540B\end{tabular}}}
\\
 & \multicolumn{1}{c}{Acc.} & \multicolumn{1}{c}{Acc.} & \multicolumn{1}{c}{\begin{tabular}[c]{@{}c@{}}Acc.\\ with Calc.\end{tabular}} & \multicolumn{1}{c}{Acc.} & \multicolumn{1}{c}{\begin{tabular}[c]{@{}c@{}}Acc.\\ with Calc.\end{tabular}} \\ \hline
\textbf{GSM8K} & 8.11 & \textbf{21.99} & 38.21 & 56.90 & 58.60 \\
\huge{Dataset Size} & \huge{6725} & \huge{5337} & \huge{5337} & \huge{-} & \huge{-} \\
\textbf{MAWPS} & 54.15 & \textbf{70.41} & 88.22 & 93.00 & 93.66 \\
\huge{Dataset Size} & \huge{1590} & \huge{1590} & \huge{1590} & \huge{-} & \huge{-} \\
\textbf{ASDiv} & 39.64 & \textbf{42.12} & 60.73 & 73.9 & 72.6 \\
\huge{Dataset Size} & \huge{1844} & \huge{1544} & \huge{1544} & \huge{-} & \huge{-} \\
\hline
\end{tabular}
}
\caption{Task accuracy across arithmetic reasoning datasets for T5 XXL without finetuning (baseline) and finetuned on PaLM 540B generated chain-of-thought (CoT). We report the accuracy of PaLM 540B on the used datasets for reference. We do not finetune PaLM for this, but employ 8 chain of thought prompts.}
\label{arithmetic_reasoning_table}
\end{table}

\subsubsection{Ablation study on generating chain-of-thought data}

We perform an ablation study to confirm that providing a LLM with the target during CoT generation is beneficial. We found that for the GSM8K dataset, PaLM 540B only achieves a 59.98\% accuracy if prompted without the target. In comparison, when including the target in the prompt the accuracy is 79.37\%. A superficial explanation would be that when the model is conditioned on the expected answer, it produces the same CoT but copies the answer. However, an analysis of a subset of the differences between CoT produced with and without this conditioning shows that most of the benefits actually come from the model correcting CoT that had a single step missing or was wrong.

\subsection{Commonsense reasoning}

For the StrategyQA dataset (Table \ref{teacher_table}), we found that using CoT finetuning improves accuracy from 68.12\% to 71.98\%, using only 1319 of the original 1648 examples. Compared to the arithmetic reasoning datasets, the improvement is not as significant. This can be explained by the model lacking factual knowledge that the dataset requires. The task is heavily focused on the model reasoning on such knowledge, however, a smaller LM is most likely not in possession of this knowledge compared to a larger model with higher memorisation capacity.

\subsection{Symbolic reasoning}

Table \ref{symbolic_reasoning_table} shows the results obtained for the synthetic symbolic reasoning datasets, focusing on OOD generalization. Focusing on Last Letter Concatenation, it can be stated that both traditional finetuning and the suggested method fail at generalizing to a longer sequence length. In comparison, the proposed method significantly increases accuracy for the Coinflip dataset with regard to generalizing to three coinflips. In contrast, generalisation to four coinflips is slightly weaker than the baseline, which performs very strongly. This may be related to the task length being twice that of the training task.

\begin{table}[ht]
\Huge
\resizebox{\columnwidth}{!}{%
\begin{tabular}{clccc}
\hline
\multicolumn{2}{l}{} &
  \textbf{\begin{tabular}[c]{@{}c@{}}Baseline\\ T5 XXL\end{tabular}} &
  \textbf{\begin{tabular}[c]{@{}c@{}}CoT Finetuned \\ T5 XXL\end{tabular}}  &
  \textbf{\begin{tabular}[c]{@{}c@{}}CoT 8-shot \\ PaLM 540B\end{tabular}} \\ \hline
\multirow{2}{*}{\textbf{\begin{tabular}[c]{@{}c@{}}Last Letter\\ Concat.\end{tabular}}} & OOD: 3          &   0.00   &   0.00 & 94.8  \\
                                                                                        & OOD: 4 &  0.00    &   0.00 & 63.0   \\
\multirow{2}{*}{\textbf{Coinflip}}                                                      & OOD: 3          & 13.10 & \textbf{86.70} & 98.6 \\
                                                                                        & OOD: 4          & \textbf{73.80} & 70.50 & 90.2 \\ \hline
\end{tabular}%
}
\caption{Task accuracy across the symbolic reasoning datasets for T5 XXL finetuned on chain-of-thought (CoT) data. For each dataset, there are 1000 training and testing examples. We report the accuracy of PaLM 540B from \citep{wei2022chain} for reference.}
\label{symbolic_reasoning_table}
\end{table}

\subsection{Replicating Results using different Teacher Models}

We demonstrate the robustness of our method using a different teacher model, namely GPT-3 175B. Table \ref{teacher_table} shows the results for GSM8K and StrategyQA when T5 XXL is finetuned on CoT data generated by GPT-3. The results show that the proposed method elicits improvements also with other LLMs as teachers. We also report the accuracy of T5 XXL finetuned on golden CoT provided with the datasets. For the StrategyQA dataset, the model finetuned on the golden CoT performs best, which may be attributed to the dataset being the largest, as both PaLM and GPT-3 get some examples wrong. In contrast, the model finetuned on PaLM generated CoT performs the best for GSM8K.

\begin{table}[ht]
\Huge
\centering
\resizebox{\columnwidth}{!}{%
\begin{tabular}{lcccccc}
\hline
\multicolumn{1}{c}{\multirow{2}{*}{}} & \multirow{2}{*}{\textbf{\begin{tabular}[c]{@{}c@{}}Base\\ Task\end{tabular}}} & \multirow{2}{*}{\textbf{\begin{tabular}[c]{@{}c@{}}Original\\ Cot\end{tabular}}} & \multicolumn{2}{c}{\textbf{\begin{tabular}[c]{@{}c@{}}CoT finetuned\\ T5 XXL using\end{tabular}}} & \multicolumn{2}{c}{\textbf{CoT 8-Shot}} \\
\multicolumn{1}{c}{} &  &  & \textbf{\begin{tabular}[c]{@{}c@{}}PaLM\\ 540B\end{tabular}} & \textbf{\begin{tabular}[c]{@{}c@{}}GPT-3\\ 175B\end{tabular}} & \textbf{\begin{tabular}[c]{@{}c@{}}PaLM\\ 540B\end{tabular}} & \multicolumn{1}{l}{\textbf{\begin{tabular}[c]{@{}l@{}}GPT-3\\ 175B\end{tabular}}} \\ \hline
\textbf{GSM8K} & 8.11 & 19.94 & \textbf{21.99} & 18.42 & 56.9 & 46.9 \\
\huge{acc. with Calc.} & \huge{-} & \huge{26.99} & \huge{38.21} & \huge{33.06} & \huge{58.6} & \huge{49.6} \\
\huge{Dataset Size} & \huge{6725} & \huge{6725} & \huge{5337} & \huge{5298} & \huge{-} & \huge{-} \\
\textbf{StrategyQA} & 68.12 & \textbf{71.98} & 67.15 & 63.77 & 77.8 & 65.4 \\
\huge{Dataset Size} & \huge{1648} & \huge{1648} & \huge{1319} & \huge{1319} & \huge{-} & \huge{-} \\ \hline
\end{tabular}%
}
\caption{Task accuracy for T5 XXL finetuned on chain-of-thought (CoT) data generated by PaLM 540B and GPT-3 175B. We also finetune on the reasoning steps provided by the datasets. We report the accuracy of PaLM 540B on the used datasets for reference. We do not finetune PaLM for this, but employ 8 chain of thought prompts.}
\label{teacher_table}
\end{table}

\subsection{Ablation study on model size}
\begin{figure}[tb]
    \centering
    \includegraphics[width=\columnwidth]{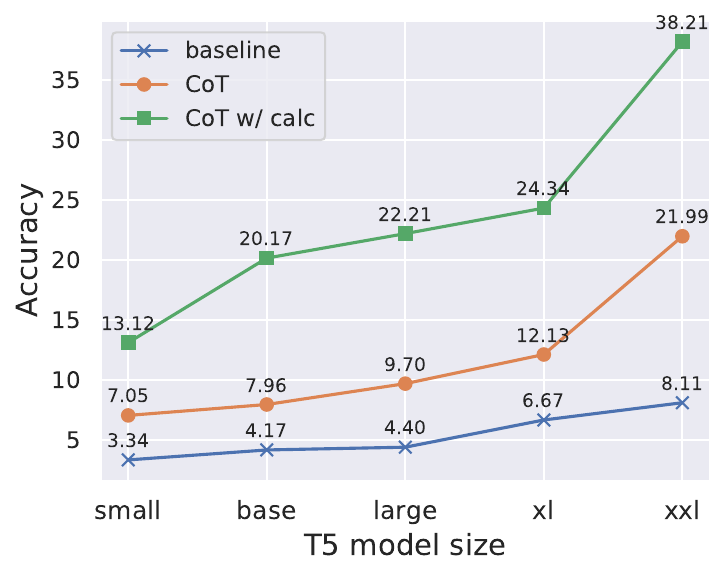}
    \caption{Effect of student model (T5) size on accuracy on GSM8K.}
    \label{fig:model_size}
\end{figure}

We investigate the performance gain achieved via finetuning student models of different sizes. Figure \ref{fig:model_size} shows the performance gain achieved when finetuning T5 of different sizes on the GSM8K dataset. Our results show that T5 base, with 44 times fewer parameters than T5 XXL, matches the performance of the baseline T5 XXL when trained on CoT data. Moreover, given an external calculator, even T5 small outperforms the baseline T5 XXL.

\subsection{Ablation study on dataset size}

\label{dataset_size_exp}
We also investigate the trade-off between the performance gain from CoT finetuning and dataset size. Table \ref{dataset_table} details the test accuracy achieved when finetuning T5 XXL on only 4\% and 20\% of the data, randomly selected. In comparison to the baseline accuracy of 8.11\% (Table \ref{teacher_table}), we see that our method is 6x more data efficient, achieving accuracy of 11.22\% with only 20\% of the examples. However, training on just 20\% of the data still creates a quality gap, and it's possible that with e.g. 200\% larger dataset we could outperform the results in Table \ref{teacher_table}.

\begin{table}[ht]
\centering
\resizebox{\columnwidth}{!}{%
\begin{tabular}{lcc}
\hline
\multirow{2}{*}{\textbf{\begin{tabular}[c]{@{}l@{}}Percentage of GSM8K\\ data used to train\end{tabular}}} & \multicolumn{2}{c}{\textbf{CoT finetuned T5 XXL}} \\
                                                                                                      & \multicolumn{1}{c}{Acc.} & \multicolumn{1}{c}{Acc. with Calc.}            \\\hline
4\% (213 examples)                                                                                    & 6.29                      & 12.28                                \\
20\% (1067 examples)                                                                                     & 11.22                     & 20.47                            \\
100\% (5337 examples)                                                                                          & 21.99                     & 38.21      \\\hline                        
\end{tabular}%
}
\caption{Task accuracy of T5 XXL finetuned on different amounts of chain-of-thought (CoT) data generated by PaLM 540B.}
\label{dataset_table}
\end{table}

\section{Discussion}

We demonstrate that finetuning larger LMs on the CoT data generated by LLMs of over 100 billion parameters can significantly improve task accuracy. Even a small number of CoT examples appear to suffice for this. However, such improvements appear to be task dependent. For example, the effects are limited for the StrategyQA dataset, which can be attributed to the task requiring specific factual knowledge, which smaller LMs may not have memorised due to their limited capacity. Nevertheless, there is some performance improvement, which may be attributed to the model learning how to approach such tasks. Moreover, the CoT knowledge distillation pipeline presented allows to trade-off model and dataset size with accuracy. Future work could explore improving the reasoning of small models in multi-task settings, as well as the generation of new training data using LLMs, rather than annotating existing datasets.

\section{Conclusion}

This work explores CoT knowledge distillation from LLMs of over 100 billion parameters to smaller LMs. We propose a knowledge distillation pipeline consisting of two keys steps: (1) generate CoT for existing datasets using LLMs and (2) finetune smaller LMs on the CoT. Our results demonstrate that finetuning on CoT improves task accuracy across a range of benchmarking datasets.

\section{Limitations}

The results we present must be viewed in the context of a few limitations. A limitation is that we only perform experiments in English and on one task at a time. To be more comparable to a LLM few-shot settings, other languages and a multi-task setup could be explored. Furthermore, in order to replicate the results access to none public models is required and inference must be performed on large amounts of data. Another limitation of our work is that it only explores the original CoT prompting approach, but we do not explore subsequent improvements, such a self-consistency \citep{wang2022self}.

\section{Ethical Considerations}

The main ethical considerations of our research arise from the text generation performed. The concerns here are that both the teacher and student model may potentially generate non-factual \citep{ji2022survey, pagnoni2021understanding, kreps2022all} or offensive output \citep{gehman2020realtoxicityprompts}. This is largely influenced by the input data, which is our case are standard, peer-reviewed benchmarking tasks in the NLP domain.

\bibliography{anthology,custom}
\bibliographystyle{acl_natbib}

\appendix

\section{Dataset Usage and Licenses}
\label{licenses}

In this section, we list the licenses for the datasets used and any ethical concerns regarding their usage. We describe the dataset splits used for all datasets in Section \ref{experimental_setup} of the paper.

\subsection{Arithmetic Reasoning}
The GSM8K dataset \citep{cobbe2021training} is available under the MIT license. The MAWPS dataset \citep{koncel2016mawps} is available under the CC BY 4.0 and the ASDiv dataset \citep{miao2021diverse} is available under the CC BY-NC 4.0 license. We follow the intended usage of the datasets.

\subsection{Commonsense Reasoning}

The StrategyQA dataset \citep{geva-etal-2021-aristotle} is available under the MIT license. Similar to \citet{wei2022chain}, we use the open-domain setting version available as part of the Big-bench collaboration \citep{bigbench}, available under the Apache License 2.0. We follow the intended usage of the datasets.

\subsection{Symbolic Reasoning}

We generate the symbolic reasoning datasets as described in \citet{wei2022chain}.

\section{Computational Resources}
\label{compute}

We perform inference and finetuning on different sizes of T5 on TPUs. We perform inference on PaLM 540B also on TPUs. Our results can be replicated via the public API (\url{https://developers.generativeai.google/products/palm}). To make requests to GPT-3 175B, we use the public API (\url{https://beta.openai.com/docs/introduction}).

\end{document}